%% file: main.tex
\documentclass[10pt,twocolumn,letterpaper]{article}

\usepackage{cvpr}      % To produce the REVIEW version
\usepackage[numbers]{natbib}

\definecolor{cvprblue}{rgb}{0.21,0.49,0.74}
\usepackage[pagebackref,breaklinks,colorlinks,allcolors=cvprblue]{hyperref}
\usepackage{multirow}
\usepackage{multirow}
\usepackage{algorithm}
\usepackage{algorithmic}
\usepackage{newfloat}
\usepackage{listings}
\usepackage{arydshln}
\usepackage{caption} % DO NOT CHANGE THIS AND DO NOT ADD ANY OPTIONS TO IT'
\usepackage{amsmath}
\usepackage{subcaption} % 加载subcaption包
\usepackage{graphicx}    % 加载graphicx包以支持插图

\def\paperID{16327} % *** Enter the Paper ID here
\def\confName{CVPR}
\def\confYear{2025}

\title{Weakly Supervised Temporal Action Localization via Dual-Prior Collaborative Learning Guided by Multimodal Large Language Models}

\author{
Quan Zhang$^{1}$\thanks{Both authors contributed equally to this research}\quad
Jinwei Fang$^{2}$\footnotemark[1]\quad
Rui Yuan$^{1}$ \quad
Xi Tang$^{1}$ \quad
Yuxin Qi$^{3}$ \quad
Ke Zhang$^{1}$\thanks{Corresponding authors.} \quad
Chun Yuan$^{1\textdagger}$\footnotemark[2]
\and
$^1$Tsinghua University \\
$^2$University of Science and Technology of China \\
$^3$Shanghai Jiao Tong University \quad
}

\begin{document}
\maketitle
\input{sec/0_abstract}    
\input{sec/1_intro}
\input{sec/2_Related_Work}

\input{sec/3_Methods}

\input{sec/Experiment}

\input{sec/Conclusion}
{
    \small
    \bibliographystyle{ieeenat_fullname}
    \bibliography{main}
}


\end{document}

%% file: sec/0_abstract.tex
\begin{abstract}
Recent breakthroughs in Multimodal Large Language Models (MLLMs) have gained significant recognition within the deep learning community, where the fusion of the Video Foundation Models (VFMs) and  Large Language Models(LLMs) has proven instrumental in constructing robust video understanding systems, effectively surmounting constraints associated with predefined visual tasks. These sophisticated MLLMs exhibit remarkable proficiency in comprehending videos, swiftly attaining unprecedented performance levels across diverse benchmarks. However, their operation demands substantial memory and computational resources, underscoring the continued importance of traditional models in video comprehension tasks. In this paper, we introduce a novel learning paradigm termed MLLM4WTAL. This paradigm harnesses the potential of MLLM to offer temporal action key semantics and complete semantic priors for conventional Weakly-supervised Temporal Action Localization (WTAL) methods. MLLM4WTAL facilitates the enhancement of WTAL by leveraging MLLM guidance. It achieves this by integrating two distinct modules: Key Semantic Matching (KSM) and Complete Semantic Reconstruction (CSR). These modules work in tandem to effectively address prevalent issues like incomplete and over-complete outcomes common in WTAL methods. Rigorous experiments are conducted to validate the efficacy of our proposed approach in augmenting the performance of various heterogeneous WTAL models. 
\end{abstract}

%% file: sec/1_intro.tex
\section{Introduction}
\label{sec:intro}

Temporal action localization (TAL) \cite{paul2018w}  aims to localize action instances of interest from untrimmed videos. Although current fully-supervised TAL methods achieve significant localization results, fully-supervised methods require expensive, time-consuming, and labor-intensive frame-level annotations, and weakly-supervised temporal action localization (WTAL) methods \cite{zhang2025rethinking}, which require only video-level labels, have received much attention in order to alleviate the cost of annotation.

\begin{figure}[t]
	\centering	\includegraphics[width=0.95\linewidth,height=3cm]{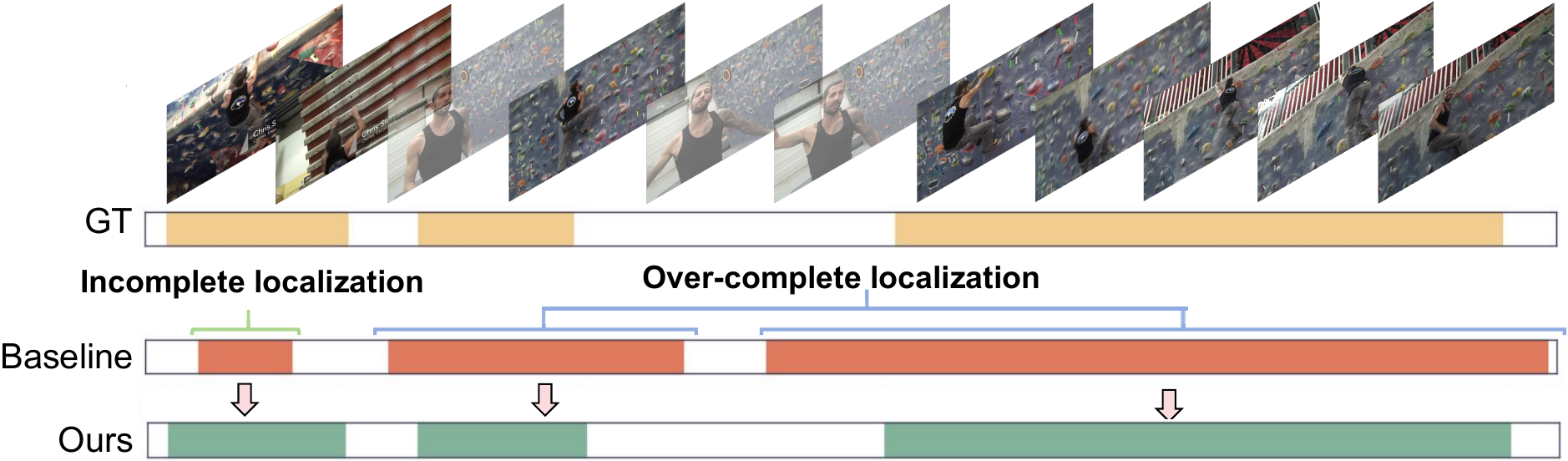}
	\caption{Previous methods have problems of over-complete positioning and incomplete positioning.}
	\label{fig:intro}
\end{figure}

Existing WTAL methods typically train a classifier\cite{zhang2025rethinking} using video-level labels, generating a set of probability predictions. This process results in temporal class activation maps (T-CAM). Despite notable performance improvements, current methods still face two challenges: incomplete and over-complete localization. As illustrated in Fig.\ref{fig:intro}, some sub-actions with low discriminability may be ignored, leading to incomplete localization, while certain background segments that contribute to classification might be misclassified as foreground actions, causing over-complete localization.

To overcome these challenges, we turn our perspective to existing multimodal large language models(MLLMs)\cite{alayrac2022flamingo}. The state-of-the-art MLLM employs a seamless amalgamation of a video foundation model and a large language model, culminating in the development of a sophisticated system for enhanced video comprehension. This integrated framework transcends the limitations inherent in traditional predefined visual tasks. Although MLLM exhibits powerful video understanding capabilities, the resource-intensive nature of running it poses high memory and computational costs. Therefore, traditional models remain essential for video understanding tasks. Is it possible to propose a learning paradigm that leverages MLLMs to enhance existing traditional WTAL methods? Inspired by this, we introduce a paradigm for MLLM-enhanced WTAL tasks, known as MLLM4WTAL.

Aiming at the incomplete and over-complete localization problems of previous WTAL methods, we propose MLLM-based key semantic matching (KSM) and complete semantic reconstruction (CSR) modules. KSM aims to utilize MLLM to generate key semantic priors for videos, matching segments in the video based on key semantic embeddings and video embeddings. This matching strategy helps locate key intervals for temporal actions. CSR aims to utilize MLLM to generate complete semantic priors for videos. By reconstructing the complete semantics of masked action words, we aim to mine the time interval of action instances as complete as possible. This reconstruction strategy helps locate complete intervals for temporal actions.

Based on KSM and CSR, we obtain the key intervals and the complete intervals of the temporal action in turn. Due to the task characteristics of KSM, it may overly focus on segments most matching key semantics, neglecting fewer matching segments, leading to incomplete localization. On the other hand, due to the task characteristics of CSR, it may associate as many segments as possible with key action verbs, often introducing background segments similar to foreground actions, resulting in over-complete localization. Considering the strengths and weaknesses of the two modules, we propose a dual prior interactive enhancement optimization strategy. First, the KSM branch performs forward operations to obtain $\mathcal{A}_{KSM}$, and then the CSR branch uses $\mathcal{A}_{KSM}$ as the optimization target to alleviate its over-complete localization problem. Subsequently, the CSR branch performs forward operations to obtain $\mathcal{A}_{CSR}$, and then the KSM branch uses $\mathcal{A}_{CSR}$ as the optimization target to alleviate its incomplete localization problem. The two modules interactively distill each other, forming a strong joint approach.

Our approach implements the state-of-the-art on two popular benchmarks, THUMOS14 and ActivityNet1.2. Furthermore, we find that our proposed approach can be applied to existing methods and improve their performance with significant advantages. Our contributions can be summarized in four aspects:
\begin{itemize}
\item We propose a paradigm, MLLM4WTAL, that guides WTAL methods using MLLM. To the best of our knowledge, this is the first work that enhances WTAL methods using MLLM. We also show that our method can be easily extended to existing state-of-the-art methods and improve their performance without introducing additional overhead in the inference phase.
\item To fully leverage the prior information provided by MLLM, we design two modules, KSM and CSR, to enhance WTAL methods from the perspectives of key semantic matching and complete semantic reconstruction, respectively.
\item To achieve a strong collaboration between KSM and CSR, we propose a dual prior interactive distillation strategy, cleverly addressing the incomplete and over-complete issues present in each module.
\item Extensive experiments show that our method outperforms existing methods on two public datasets. Comprehensive ablation studies reveal the effectiveness of the proposed method.
\end{itemize}

%% file: sec/2_Related_Work.tex
\section{Related Work}
\label{sec:formatting}
\noindent\textbf{Temporal Action Localizationl} requires the simultaneous prediction of action start and end times along with the corresponding category. In fully-supervised temporal action localization training, frame-level labeling is commonly employed. Inspired by the widespread application of deep learning techniques\cite{zhang2024distilling,zhang2025imdprompter,chen2024gim,wei2024free,wei2025modeling,wei2024task}, existing methods mainly adopt a deep learning-based technical approach, which can be divided into the following two categories:  top-down and bottom-up. Top-down methods \cite{2018Rethinking} transform action localization into a time-dimensional detection task, inspired by image object detection. These methods generate action proposals, followed by classification and regression on temporal boundaries. They can draw on state-of-the-art techniques for Object Detection. On the other hand, bottom-up methods\cite{xu2020g} generate frame-level predictions, followed by action localization using cleverly designed post-processing techniques. However, this fully supervised approach relies heavily on frame-by-frame annotation, which is both expensive and impractical when dealing with long videos. 
% \subsection{todo:need a subtitle}
% \subsection{Key Semantic Matching}\label{KSM}

\noindent\textbf{Weakly-Supervised Temporal Action Localization} has garnered increasing attention within the research community, particularly due to its ability to learn from video-level category labels during training, thus circumventing the need for precise temporal annotations\cite{paul2018w}. The initial approach was UntrimmedNet \cite{wang2017untrimmednets}, which first addressed this problem by employing a Multiple Instance Learning (MIL) framework that selects foreground segments and combines them into action segments. Subsequent improvement methods, such as STPN \cite{nguyen2018weakly}, enhanced UntrimmedNet by introducing a sparse loss to make the selected segments more sparse. CoLA \cite{zhang2021cola} utilizes contrastive learning to distinguish between foreground and background segments. In addition, UGCT \cite{yang2021uncertainty} proposed an online pseudo-label generation mechanism with uncertainty-aware learning capability for pseudo-label supervision of attention weights. 
% \par
\begin{figure*}[t]
	\centering
	\includegraphics[width=0.95\linewidth]{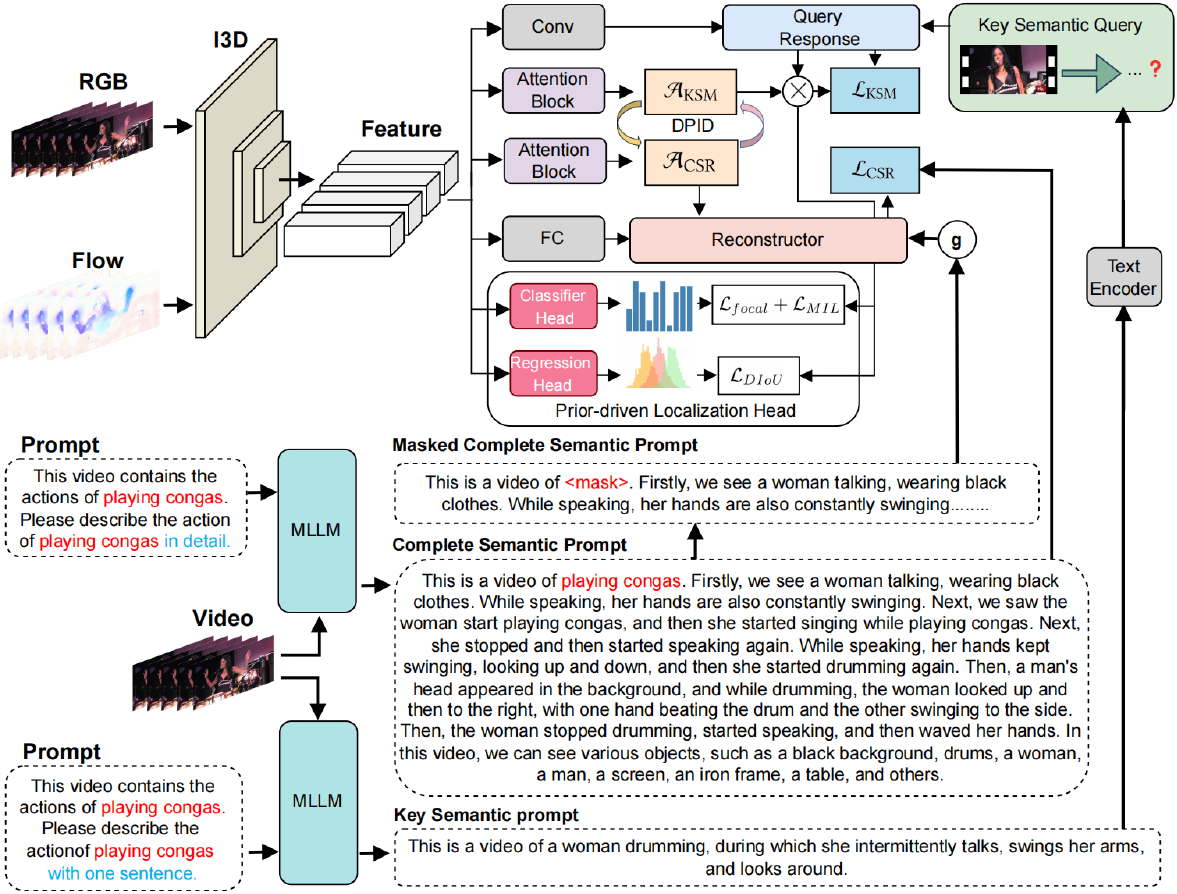}
	\caption{The overview of our method. In this work, Key Semantic Matching aims to mine critical temporal intervals of temporal actions in videos with the help of matching key semantics provided by MLLM with video segments. 
Complete Semantic Reconstruction aims to utilize the complete semantics provided by MLLM to reconstruct the complete semantics of masking key action words, thus mining as complete an action instance range as possible. The dual prior interactive distillation(DPID) between $\mathcal{A}_{KSM}$ and $\mathcal{A}_{CSR}$ achieves a powerful combination of Key Semantic Matching and Complete Semantic Reconstruction.Prior-driven Localization Head aims to refine pseudo-labels while avoiding the high computational overhead of using MLLM during inference.}
	\label{fig:pipe}
\end{figure*}
% \subsection{todo:need a subtitle}

\noindent\textbf{Multimodal Large Language Models} have emerged as a powerful paradigm for tackling a wide range of tasks spanning multiple modalities. LLM \cite{brown2020language} has recently excelled in Natural Language Processing (NLP) tasks. To integrate with other modalities, researchers have worked to construct MLLM \cite{alayrac2022flamingo,li2023blip}. Flamingo \cite{alayrac2022flamingo} integrates pre-trained visual and linguistic models to achieve state-of-the-art performance using a small number of samples. BLIP-2 \cite{li2023blip} proposes a generic and efficient pre-training strategy that utilizes frozen pre-trained image encoders and large linguistic models for visual language pre-training. MiniGPT4 \cite{zhu2023minigpt} implements the system by aligning frozen visual coders and LLMs, Vicuna \cite{chiang2023vicuna}, using only one projection layer. Otter \cite{li2023otter} demonstrates improved instruction-following and context-learning capabilities. In the domain of video, ChatVideo \cite{wang2023chatvideo} implements user interaction with LLMs using the tracklet as the fundamental video unit. As for Video-LLaMA \cite{zhang2023video}, it utilizes the pre-trained models ImageBind \cite{girdhar2023imagebind} and LLaMA \cite{touvron2023llama} on top of BLIP-2, which further facilitates cross-modal training for video.

%% file: sec/3_Methods.tex
\section{Methods}
% \subsection{todo:is this a subsection? Problem definition}
\noindent\textbf{Problem definition.} Given an untrimmed video containing action instances from C categories, WTAL aims to predict a set of action instances $\left\{\phi \mid \phi_{\mathrm{i}}=\left(\mathrm{c}, \mathrm{q}, \mathrm{t}_{\mathrm{s}}, \mathrm{t}_{\mathrm{e}}\right)\right\}$ using only video-level labels $\mathrm{y} \in\{0,1\}^{\mathrm{c}}$. For each action instance, $\mathrm{c}$ represents the predicted category, $\mathrm{q}$ represents the confidence, and $\mathrm{t}_{\mathrm{s}}$ and $\mathrm{t}_{\mathrm{e}}$ represent the start timestamp and end timestamp, respectively.

\noindent\textbf{Overview.} 
The proposed MLLM4WTAL framework, illustrated in Fig.\ref{fig:pipe}, enhances WTAL through two key aspects: Key Semantic Matching (KSM) and Complete Semantic Reconstruction (CSR), leveraging the semantic priors of MLLM. A dual priors interactive enhancement strategy is employed for mutual enhancement of these priors. 

In section \ref{KSM}, MLLM first processes the raw video input to extract a temporal action key description. A key semantic query prompt is then constructed and input into the text encoder to obtain the key semantic query. RGB and flow video features $X_{rgb}$ and $X_{flow}$ are passed into the video embedding module to generate video feature embeddings. The KSM module matches the key semantic query with video segments, generating a contrastive query response that activates the most relevant video segment. Attention weights are applied to suppress background segments' responses.

In section \ref{CSR}, MLLM inputs the raw video to generate a complete temporal action description and masks key action words in the sentence. An attention mechanism gathers fragments related to the masked words, and a linguistic reconstructor predicts the masked words.

Finally, section \ref{DPID} utilizes a dual priors interactive enhancement strategy to foster collaboration between KSM and CSR, improving the accuracy and completeness of the localization results.

\subsection{Key Semantic Matching}\label{KSM}
In this section, we will introduce key semantic matching (KSM) to comprehensively activate key temporal action intervals. Specifically, KSM consists of a key semantic generator, video embedding module, text embedding module, and semantic matching module.

\noindent\textbf{\bfseries Key Semantic Generator.} In the key semantic generator module, the first step is to construct key semantic prompts for MLLM, guiding the MLLM to generate key descriptions for temporal actions. It is worth noting that when building the prompts, we have prior knowledge of the action categories present in the video. We introduce this action category before our prompt template. The prompt template $P_{\text {key}}$ format is as follows:

\textbf{This video contains the actions of $\{Cls\}$. Please describe the action of $\{Cls\}$ with one sentence.} 

\noindent Input the prompt template $P_{\text {key }} $ and the original video V into MLLM to get the key semantic description of the video $D_{\text {key }}$ as:
\begin{equation}
D_{\text {key }}=\Phi_{\text {MLLM }}\left(P_{\text {key }}, V\right)
\end{equation}
\textbf{\bfseries Video Embedding Module}. Similar to other WTAL models, the video embedding module consists of two 1D convolutions followed by ReLU and Dropout layers. We adopt the same strategy as in \cite{bai2020boundary} to fuse the RGB and flow features to obtain the input $\mathrm{F} \in \mathrm{R}^{\mathrm{T} \times 2048}$ of the video embedding module. 
\begin{equation}
F=\Phi_{\text {fuse }}\left(F_{\text {rgb }}, F_{\text {flow }}\right) 
\end{equation}
The corresponding video feature embedding $\mathrm{F}_{\mathrm{e}} \in \mathrm{R}^{\mathrm{T} \times 2048}$ can then be obtained by $ F_e=\Phi_{\text {emb}}(F)$, where $\Phi_{\text{emb}}$ represents the video embedding module. In addition, following previous works \cite{hong2021cross,islam2021hybrid}, an attention mechanism is used to generate attention weight $\mathcal{A}_{KSM} \in \mathrm{R}^{\mathrm{T} \times 1}$ for each video clip.
\begin{equation}
\mathcal{A}_{K S M}=\operatorname{sigmoid}\left(\Phi_{\text {attention}}\left(F_e\right)\right)
\end{equation}
where $\Phi_{\text {attention}}$ is an attention mechanism containing some convolutional layers.

\noindent\textbf{\bfseries Text Embedding Module}. The text embedding module aims to generate key semantic queries using key semantic descriptions. We have used the learnable prompt to obtain the input $T_{\text {query }}$ for the text embedding module.
\begin{equation}
T_{\text {key}}=\Phi_{\text {GloVe}}\left(D_{\text {key}}\right) \\
\end{equation}
\begin{equation}
T_{\text {query}}=\left[T_{\text {start}}, T_{\text {context}}, T_{\text {key}}\right],
\end{equation}
where $T_{\text{start}}$ represents the [START] token that was randomly initialized, $T_{\text{context}}$  is the learnable textual context of length $N_{context}$, and $T_{\text{key}}$ is the key informative textual feature embedded with GloVe \cite{openai-gpt4}. Additionally, the $(C+1)th$ additional context class embedding is initialized with 0.
Subsequently, a transformer encoder, denoted as $\Phi_{\text{trans}}$, functions as the text embedding module to generate text queries. Importantly, it should be noted that the key semantic query $F_{\text{query}}$ can be obtained through $F_{\text{query}}=\Phi_{\text{trans }}\left(T_{\text{query}}\right)$, where $F_{\text{query}} \in \mathrm{R}^{(\mathrm{C+1}) \times 2048}$.

\noindent\textbf{\bfseries Semantic Matching Module.} The semantic matching module aims to activate video segment features associated with key semantic queries. Specifically, we perform inner product operation on video embedding feature $X_v$ and text embedding feature $X_t$ to generate segment-level video-text similarity matrix $M \in R^{T \times(C+1)}$. In addition to this, following \cite{hong2021cross}, we also employ the attention weight $\mathcal{A}_{KSM}$ to suppress the background segment responses. The segment-level matching result $\hat{M} \in R^{T \times(C+1)}$ for background suppression can be obtained by $\hat{M}=\mathcal{A}_{K S M} * M$, where $'*'$ denotes the element-wise multiplication.
Following \cite{paul2018w}, we employ top-k multi-instance learning to compute the matching loss. Specifically, we compute the average of top-k similarities on the time dimension of the response of a particular text query category as the video-level video-text similarity.
For the j-th action category, the video-level similarities $\mathrm{S}_{\mathrm{j}}$ and $\hat{S}_{\mathrm{j}}$ are generated from $M$ and $\hat{M}$, respectively.
\begin{equation}
s_j=\max _{\substack{l \subset\{1, \ldots, \mathcal{T}\} \\
|l|=k}} \frac{1}{k} \sum_{i \in l} M_i(j) \\
\end{equation}
\begin{equation}
\hat{s}_j=\max _{\substack{l \subset\{1, \ldots, \mathcal{T}\} \\
|l|=k}} \frac{1}{k} \sum_{i \in l} \hat{M}_i(j) \\
\end{equation}

Where l is a set of indexes containing the top k segments with the highest similarity to the j-th text query, and k is the number of selected segments.

We then apply softmax to $\mathrm{S}_{\mathrm{j}}$ and $\hat{S}_{\mathrm{j}}$ to generate video-level similarity scores $\mathrm{p}_{\mathrm{j}}$ and $\hat{p}_{\mathrm{j}}$.

\begin{equation}
p_j=\operatorname{softmax}\left(s_j\right) \\
\end{equation}
\begin{equation}
\hat{p}_j=\operatorname{softmax}\left(\hat{s}_j\right) \\
\end{equation}

We encourage positive scores for video-text category matching to approach 1, while negative scores for training KSM objectives should be close to zero.

Here, $\mathrm{y}_{\mathrm{j}}$ and $\hat{y}_{\mathrm{j}}$ represent the video-text matching labels. In $\mathrm{y}_{\mathrm{j}}$, the additional $(C+1)$-th background class is labeled as 1, while in $\hat{y}_{\mathrm{j}}$, it is labeled as 0. The $L_{K S M}$ expression is as follows:
\begin{equation}
L_{K S M}=-\sum_{j=1}^{C+1}\left(y_j \log \left(p_j\right)+\hat{y}_j \log \left(\hat{p}_j\right)\right)
\end{equation}

\subsection{Complete Semantic Reconstruction}\label{CSR}
Complete Semantic Reconstruction aims to complete the masked keywords in video descriptions by focusing on as comprehensive as possible text-related video clips. The proposed CSR also contains a video embedding module and a text embedding module, in addition to the transformer reconstructor for multimodal interaction and reconstruction of the complete text description.

\noindent\textbf{\bfseries Complete Semantic Generator}. In the Complete Semantic Generator module, we first need to construct MLLM complete semantic prompts that prompt the MLLM to generate complete information of action instance. It is worth noting that when constructing the cue, we know the action category that the video contains, and we introduce this action category a priori into our prompt cue template, which is formatted as follows: 

\textbf{This video contains the actions of ${Cls}$. Please describe the action of ${Cls}$ in detail.}

\noindent Enter the cue template $P_{\text {complete}}$ and original video $V$ into MLLM to get a description of the complete information of the video $D_{complete}$
\begin{equation}
D_{\text {complete }}=\Phi_{\text {MLLM }}\left(P_{\text {complete }}, V\right)
\end{equation}

\noindent\textbf{\bfseries Video Embedding Module}. Given the original video features $\mathrm{F} \in \mathrm{R}^{\mathrm{T} \times 2048}$, we obtain the corresponding video feature embedding 
$\mathrm{F}_{\text{complete}} \in \mathrm{R}^{\mathrm{T} \times 512}$ through a fully connected layer. 
\begin{equation}
F_{\text {complete }}=\Phi_{\mathrm{FC}}(F) 
\end{equation}
To explore the complete temporal intervals of video that are semantically related to text, CSR includes an attention mechanism similar to \ref{KSM}. The attention weights $\mathcal{A}_{CSR}$ of the CSR module can be obtained through the following equation: 
\begin{equation}
\mathcal{A}_{C S R}=\operatorname{sigmoid}\left(\Phi_{\text{attention}}\left(F_e\right)\right)
\end{equation}
where $\Phi_{\text{attention}}$ is an attention mechanism containing some convolutional layers.

\noindent\textbf{\bfseries Text Embedding Module}. We mask the key action words of the video-complete description and embed the masked sentences with GloVe\cite{pennington2014glove} and a fully connected layer to obtain the complete semantic embedding $\mathrm{F}_{\mathrm{c}} \in \mathrm{R}^{\mathrm{m} \times 512}$, where m is the length of the sentence.

\noindent\textbf{\bfseries Transformer Reconstructor}. The Transformer reconstructor is used to reconstruct masked description sentences in the Complete Semantic Reconstruction module. First, Following \cite{zhai2020two}, we randomly mask 1/3 of the words in the sentence as alternative description sentences, which may lead to a high probability of masking action labeling text. Then, the transformer's encoder is utilized to obtain foreground video features $\mathrm{F}_{fg} \in \mathrm{R}^{\mathrm{T} \times 512}$

\begin{equation}
\begin{aligned}
& \mathbf{F}_{\mathrm{fg}} =\Phi_{\text {enc }}\left(\mathbf{F}_{\text {complete }}, \mathcal{A}_{\mathrm{CSR}}\right)\\
& =\delta\left(\frac{\mathbf{F}_{\text {complete }} \mathbf{W}_q\left(\mathbf{F}_{\text {complete }} \mathbf{W}_k\right)^T}{\sqrt{d_h}} * \mathcal{A}_{\mathrm{CSR}}\right) \mathbf{F}_{\text {complete }} \mathbf{W}_v
\end{aligned}
\end{equation}
Where $\Phi_{\text {enc}}$ is the Transformer encoder, $\delta(\cdot)$ is the softmax function, $W_q, W_k, W_v \in \mathrm{R}^{512 \times 512}$ are the learnable parameters, and $d_h = 512$ is the feature dimension of $\mathbf{F}_{\text {complete }}$.

The Transformer decoder is employed to obtain a multimodal representation 
 $\mathrm{H} \in \mathrm{R}^{\mathrm{m} \times 512}$ to reconstruct masked sentences.
\begin{equation}
\begin{aligned}
\mathbf{H} & =\Phi_{\mathrm{dec}}\left(\mathbf{F}_c, \mathbf{F}_{\mathrm{fg}}, \mathcal{A}_{\mathrm{CSR}}\right) \\
& =\delta\left(\frac{\mathbf{F}_c \mathbf{W}_{q d}\left(\mathbf{F}_{\mathrm{fg}} \mathbf{W}_{k d}\right)^T}{\sqrt{d_h}} * \mathcal{A}_{\mathrm{CSR}}\right) \mathbf{F}_{\mathrm{fg}} \mathbf{W}_{v d},
\end{aligned}
\end{equation}
Where $\Phi_{\mathrm{dec}}$ is the Transformer encoder, $\delta$ is the softmax function, $W_{qd}, W_{kd}, W_{vd} \in \mathrm{R}^{512 \times 512}$ are the learnable parameters.

Finally, the probability distribution of the $i_{th}$ word $w_i$ can be obtained by the following equation:
\begin{equation}
\mathbf{P}\left(w_i \mid \mathbf{F}_{\text {complete}}, \mathbf{F}_{c[0: i-1]}\right)=\operatorname{softmax}\left(\Phi_{\mathrm{FC}}(\mathbf{H})\right)
\end{equation}
Where $\Phi_{\mathrm{FC}}$ is the fully connected layer .

The final $\mathcal{L}_{\text {CSR}}$ loss function is as follows:
\begin{equation}
\mathcal{L}_{\text {CSR }}=-\sum_{i=1}^m \log \mathbf{P}\left(w_i \mid \mathbf{F}_{\text {complete}}, \mathbf{F}_{\text {txt }[0: i-1]}\right)
\end{equation}

\subsection{Dual Prior Interactive Distillation}\label{DPID}
Following PivoTAL\cite{Nayeem_Mittal_Yu_Hall_Sajeev_Shah_Chen_Microsoft}, we utilized a prior-guided localization head. The loss function for the localization head is as follows:
\begin{equation}
    \mathcal{L}_{l o c}=\mathcal{L}_{\text {focal }}+\mathcal{L}_{D I o U}+\mathcal{L}_{M I L} .
\end{equation}

The matching strategy employed in KSM tends to focus on better matching video segments that are crucial for describing the text, resulting in high true negatives but suffering from significant false negatives. On the other hand, CSR tends to focus on all video segments relevant to the complete text description to achieve complete description reconstruction, leading to high true positives but facing serious false positive issues. To address these challenges, an interactive distillation strategy is proposed to achieve a strong joint collaboration between the two branches.

The optimization process of the whole framework is divided into two phases, Key Semantic Matching and Complete Semantic Reconstruction, and the two phases are carried out alternately:

The loss function for the Key Semantic Matching phase is as follows:
\begin{equation}
\mathcal{L}_{\text {match }}=L_{\mathrm{KSM}}+\lambda_1 \operatorname{MSE}\left(\mathcal{A}_{\mathrm{KSM}}, \psi\left(\mathcal{A}_{\mathrm{CSR}}\right)\right)+\mu_1\mathcal{L}_\text{loc}
\end{equation}

The loss function for the Complete Semantic Reconstruction phase is as follows:
\begin{equation}
\mathcal{L}_{\text {rec }}=\mathcal{L}_{\mathrm{CSR}}+\lambda_2 \operatorname{MSE}\left(\mathcal{A}_{\mathrm{CSR}}, \psi\left(\mathcal{A}_{\mathrm{KSM}}\right)\right)
\end{equation}

Interactive distillation optimization can encourage $\mathcal{A}_{KSM}$ trained by KSM and $\mathcal{A}_{CSR}$ trained by CSR to focus on the same action regions in the video. Through this strategy, it is possible to alleviate localization errors caused by the matching strategy's excessive attention to the most relevant segments. In addition, the information of complete text description can be transferred from the Complete Semantic Reconstruction model to the WTAL model through the interactive distillation strategy.

\subsection{Model Inference} 
During the training phase, we used MLLM to provide key semantics and complete semantic priors. However, during the inference phase, MLLM is no longer required, and only the prior-guided localization head needs to be executed, thus avoiding the significant computational overhead that could result from running MLLM

% \begin{figure*}[t]
% 	\centering
% 	\includegraphics[width=0.8\linewidth]{sec/fig/vis2.pdf}
% 	\label{fig:qua}
% \end{figure*}
% \begin{figure*}[t]
% 	\centering
% 	\includegraphics[width=0.8\linewidth]{sec/fig/vis3.pdf}
% 	\caption{Visualization of ground-truth and predictions.}
% 	\label{fig:qua}
% \end{figure*}

% \begin{figure*}[t]
% 	\centering
% 	\includegraphics[width=0.8\linewidth]{sec/fig/vis1.pdf}
% 	\label{fig:qua}
% \end{figure*}
% \begin{figure*}[t]
% 	\centering
% 	\includegraphics[width=0.8\linewidth]{sec/fig/vis2.pdf}
% 	\label{fig:qua}
% \end{figure*}
% \begin{figure*}[t]
% 	\centering
% 	\includegraphics[width=0.8\linewidth]{sec/fig/vis3.pdf}
% 	\caption{Visualization of ground-truth and predictions.}
% 	\label{fig:qua}
% \end{figure*}

%% file: sec/Experiment.tex
\section{Experiment}

\begin{table*}[!ht]
    \centering
    % \vspace{-3mm}
    
    \renewcommand{\arraystretch}{1.2}  % Adjust the row height
    \resizebox{0.98\textwidth}{!}{
    \begin{tabular}{c|c|c|c c c c c c c|c c c}
    \hline
        \multirow{2}{*}{Sup} & \multirow{2}{*}{Method} & \multirow{2}{*}{Pub} & \multicolumn{7}{c|}{mAP@IoU(\%)} & \multicolumn{3}{c}{AVG}\\ \cline{4-13}
        ~ & ~ & ~ & 0.1 & 0.2 & 0.3 & 0.4 & 0.5 & 0.6 & 0.7 & 0.1:0.5 & 0.3:0.7 & 0.1:0.7 \\ \hline
        \multirow{2}{*}{Full} & SSN\cite{zhao2017temporal} & ICCV2017 & 66.0 & 59.4 & 51.9 & 41 & 29.8 & - & - & 55.7 & - & - \\ 
        ~ & P-GCN\cite{zeng2019graph} & CVPR2019 & 69.5 & 67.8 & 63.6 & 57.8 & 49.1 & - & - & 61.6 & - & - \\ \hline
        \multirow{15}{*}{Weak} & CoLA\cite{zhang2021cola} & CVPR2021 & 66.2 & 59.5 & 51.5 & 41.9 & 32.2 & 22.0 & 13.1 & 50.3 & 32.1 & 40.9 \\ 
        ~ & DCC\cite{li2022exploring} & CVPR2022 & 69.0 & 63.8 & 55.9 & 45.9 & 35.7 & 24.3 & 13.7 & 54.1 & 35.1 & 44.0 \\ 
        ~ & Li et al.\cite{li2023boosting} & CVPR2023 & 71.1 & 65.0 & 56.2 & 47.8 & 39.3 & 27.5 & 15.2 & 55.9 & 37.2 & 46.0 \\ 
       ~ & Wang et al.\cite{Wang_Li_Wang} &CVPR2023 & 73.0 & 68.2 & 60.0 & 47.9 & 37.1 & 24.4 & 12.7 & 57.2 & 36.4 & 46.2 \\ 
       ~ & Ren et al.\cite{77} &CVPR2023 & 71.8 & 67.5 & 58.9 & 49.0 & 40.0 & 27.1 & 15.1 & 57.4 & 38.0 & 47.1 \\ 
       ~ & Ju et al.\cite{ju2023distilling} &CVPR2023 & 73.5 & 68.8 & 61.5 & \textbf{53.8} & 42.0 & 29.4 & 16.8 & 59.9 & 40.7 & 49.4 \\ 
       ~ & PivoTAL\cite{Nayeem_Mittal_Yu_Hall_Sajeev_Shah_Chen_Microsoft} &CVPR2023 & 74.1 & 69.6 & 61.7 & 52.1 & 42.8 & 30.6 & 16.7 & 60.1 & 40.8 & 49.7 \\ 
       ~ & AHLM\cite{AHLM} &ICCV2023 & \textbf{75.1} & 68.9 & 60.2 & 48.9 & 38.3 & 26.8 & 14.7 & 58.3 & 37.8 & 47.6 \\
        ~ & CASE\cite{Liu_Wang_Rong_Li_Zhang} &ICCV2023 & 72.3 & 67.1 & 59.2 & 49.4 & 37.7 & 24.2 & 13.7 & 57.1 & 36.8 & 46.2 \\ 
        ~ & ISSF\cite{Yun_Qi_Wang_Ma_2023} &AAAI2024 & 72.4 & 66.9 & 58.4 & 49.7 & 41.8 & 25.5 & 12.8 & 57.8 & 37.6 & 46.8 \\ 
        \cline{2-13}
        ~ & DELU\cite{chen2022dual} & ECCV2022 & 71.5 & 66.2 & 56.5 & 47.7 & 40.5 & 27.2 & 15.3 & 56.5 & 37.4 & 46.4 \\
        ~ & \textbf{MLLM4WTAL (DELU)} & - & 72.3 & 67.0 & 57.3 & 48.2 & 42.4 & 29.7 & \textbf{18.4} & 57.4 & 39.2 & 47.9 \\ 
        ~ & Zhou et al.\cite{zhou2023improving} & CVPR2023 & 74.0 & 69.4 & 60.7 & 51.8 & 42.7 & 26.2 & 13.1 & 59.7 & 38.9 & 48.3 \\ 
        ~ & \textbf{MLLM4WTAL (Zhou et al.)} & -& 74.3 & \textbf{69.8} & \textbf{61.8} & 52.3 & \textbf{43.0} & \textbf{30.8} & 16.6 & \textbf{60.2} & \textbf{40.9} & \textbf{49.8}
\\ \hline
    \end{tabular}
    }
    
    \caption{Experimental results of different methods in THUMOS14 dataset}
    \label{table1}
\end{table*}
\begin{table}[!ht]
    \centering
    % \vspace{-3mm}
    \renewcommand{\arraystretch}{1.2} 
    \resizebox{0.47\textwidth}{!}{
    \begin{tabular}{c|c|c c c|c}
    \hline
        \multirow{2}{*}{Method} & \multirow{2}{*}{Pub} &\multicolumn{3}{c|}{mAP(\%)@IoU} &\multirow{2}{*}{Avg} \\ \cline{3-5}
        ~ & ~& 0.5 & 0.75 & 0.95 &  \\ \hline
        CoLA\cite{zhang2021cola} & CVPR2021& 42.7 & 25.7 & 5.8 & 26.1 \\ 
        ACGNet\cite{lee2021weakly} & AAAI2022& 41.8 & 26.0 & 5.9 & 26.1 \\ 
        DGCNN\cite{shi2022dynamic} & MM2022& 42.0 & 25.8 & 6.0 & 26.2 \\ 
        P-MIL\cite{77} & CVPR2023	&44.2 & 26.1 &	5.3	&26.5\\ 
        ECM\cite{zhao2022equivalent} & TPAMI2023 & 41.0 & 24.9 & 6.5 & 25.5 \\
        Ren et al.\cite{77} &CVPR2023 & 44.2 & 26.1 & 5.3 & 25.5 \\ 
        Ju et al.\cite{ju2023distilling} &CVPR2023 & 48.3 & 29.3 & 6.1 & 29.6 \\ 
        CASE\cite{Liu_Wang_Rong_Li_Zhang} &ICCV2023 & 43.8 & 27.2 & 6.7 & 27.9 \\ 
        \hline
        DELU\cite{chen2022dual} & ECCV2022& 44.2 & 26.7 & 5.4 & 26.8 \\
        \textbf{MLLM4WTAL (DELU)} &-& 45.0 & 28.1 & 6.1 & 28.9 \\ 
        DDG-Net\cite{Tang_Fan_Luo_Zhang_Zhang_Yang} &ICCV2023 & 44.3 & 26.9 & 5.5 & 27.0 \\ 
        \textbf{MLLM4WTAL (DDG-Net)} &-& \textbf{48.3} & \textbf{30.1} & \textbf{6.8} & \textbf{30.1} \\ \hline
    \end{tabular}
    }
    \caption{Experimental results of different methods in ActivityNet-v1.2 dataset}
     \label{table2}
\end{table}
\begin{table}[!ht]
    \centering
    % \vspace{-3mm}
    \renewcommand{\arraystretch}{1.2} 
    \resizebox{0.45\textwidth}{!}{
    \begin{tabular}{c|c c c c c|c}
    \hline
        \multirow{2}{*}{Method} & \multicolumn{5}{c|}{mAP(\%)@IoU} &\multirow{2}{*}{Avg} \\ \cline{2-6}
         ~ & 0.3 & 0.4 & 0.5 & 0.6 & 0.7 & ~   \\ \hline
        Baseline & 56.5 & 47.7 & 40.5 & 27.2 & 15.3 & 37.4 \\ 
        Baseline+KSM & 56.8 & 47.9 & 41.9 & 28.8 & 17.7 & 38.9  \\ 
        Baseline+CSR & 56.9 & 47.8 & 41.7 & 28.6 & 17.5 & 38.7 \\ 
        Baseline+KSM+CSR & 57.3 & 48.2 & 42.4 & 29.7 & 18.4 & 39.2 \\  \hline
    \end{tabular}
    }
    \caption{Evaluation of each component of our method.(Unless otherwise specified, the THUMOS14 dataset is assumed by default.)}
    \label{table3}
\end{table}
\begin{table}[!ht]
    \centering
    % \vspace{-3mm}
    \renewcommand{\arraystretch}{1.2} 
    \resizebox{0.45\textwidth}{!}{
    \begin{tabular}{c|c|c|c c c|c}
    \hline
        \multirow{2}{*}{Method} &\multirow{2}{*}{KSM} & \multirow{2}{*}{CSR} & \multicolumn{3}{c|}{mAP(\%)@IoU} &\multirow{2}{*}{Avg} \\ \cline{4-6}
         ~ & ~ & ~ & 0.3 & 0.5 & 0.7 & ~   \\ \hline
        \multirow{3}{*}{UM\cite{lee2021weakly}} & ~ & ~ & 51.2 & 30.4 & 10.2 & 30.3 \\ 
        ~ & \checkmark & ~ & 52.4 & 33.0 & 12.5 & 32.1 \\ 
        ~ & \checkmark & \checkmark & 52.8 & 33.6 & 13.5 & 32.6 \\ \hline
        \multirow{3}{*}{CO2-Net\cite{hong2021cross}} & ~ & ~ & 54.8 & 38.4 & 13.8 & 36.0 \\ 
        ~ & \checkmark & ~ & 55.9 & 40.8 & 16.6 & 37.9 \\ 
        ~ & \checkmark & \checkmark & 56.2 & 41.3 & 17.5 & 38.2 \\ \hline
    \end{tabular}
    }
    \caption{Generic validation of our method}
    \label{table4}
\end{table}

\begin{table}[!h]
\centering
\tiny
\setlength{\abovecaptionskip}{0cm}
\setlength{\belowcaptionskip}{-0.2cm}
\vspace{-1.5em}
% \caption{todo:title}
\label{table--}
\renewcommand{\arraystretch}{1}
\resizebox{\linewidth}{!}{
    \begin{tabular}{ccccccc}
    \toprule
                             & \multicolumn{5}{c}{{\color[HTML]{000000} { mAP@IoU(\%)}}} &                       \\ \cline{2-6}
    \multirow{-2}{*}{Method} & 0.3        & 0.4        & 0.5        & 0.6       & 0.7       & \multirow{-2}{*}{AVG} \\
    \midrule
    Class name               & 56.7       & 47.9       & 40.4       & 27.6      & 15.9      & 37.7                  \\
    key semantic             & 57.3       & 48.2       & 42.4       & 29.7      & 18.4      & 39.2 {\color[HTML]{FE0000}(+1.5)}                 \\
    \hline
    a video of the {[}CLS{]} & 56.4       & 48.0       & 40.5       & 27.5      & 15.8      & 37.6                  \\
    complete semantic        & 57.3       & 48.2       & 42.4       & 29.7      & 18.4      & 39.2 {\color[HTML]{FE0000}(+1.6)} \\
    \bottomrule
\end{tabular}%
    }
\caption{Ablation study on prior information in MLLM}
\label{ablation-mllm}
\end{table}

In this section, we first show the dataset and implementation details and then compare our approach with recent state-of-the-art methods. Next, we perform a series of ablation studies to validate the effectiveness of our proposed components. In addition to this, we perform a generalization validation of our MLLM4WTAL framework. Finally, qualitative analyses are presented at the end of this section.

% \subsection{Datasets}
\noindent\textbf{\bfseries 
THUMOS14\cite{idrees2017thumos}} contains 413 untrimmed videos covering 20 different categories of actions. On average, each video contains about 15 unique instances of actions. We trained using 200 validation videos and evaluated on 213 test videos. Despite the small size of the dataset, it is challenging because the videos vary greatly in length and action instances occur more frequently in each video.

\noindent\textbf{\bfseries ActivityNet-v1.2\cite{caba2015activitynet}} covers 9682 videos that include 100 different classes of actions. These videos were categorized into 4619 training videos, 2383 validation videos, and 2480 test videos. Almost all of the videos contain at least one action category, and the action duration usually occupies more than half of the video duration. We performed model training on the training set and performance evaluation on the validation set.
% \subsection{Implementation Details}

\noindent\textbf{\bfseries Evaluation metrics.} Following the standard protocol for temporal action localization [6], we chose the mean accuracy at different intersection-union ratio (IoU) thresholds (mAP@IoU) to evaluate the model performance. For a fair comparison, we use an IoU threshold range of [0.1:0.1:0.7] on the THUMOS-14 dataset and an IoU threshold range of [0.5:0.05:0.95] on ActivityNet-v1.2. The official evaluation code provided by ActivityNet is used to assess the model's performance.

% \noindent\textbf{\bfseries Feature extraction.} Building upon our prior studies \cite{tan2021relaxed,min2020adversarial,ding2021kfc,narayan20193c}, we initially partitioned each video into 16 non-overlapping frames and applied the TV-L1 algorithm \cite{nguyen2018weakly} to compute optical flow. Following this, the number of sampled segments for the THUMOS-14 and ActivityNet-v1.2 datasets was configured at 750 and 50, respectively. Subsequently, we employed the I3D network \cite{carreira2017quo}, pre-trained on the Kinetics400\cite{kay2017kinetics} dataset, to generate RGB and optical flow features. Ultimately, the RGB and optical flow features were concatenated, resulting in extracted features with a dimensionality of 2048.

\noindent\textbf{\bfseries Training Settings.} We employed the Adam optimizer with a learning rate of 0.0005 and a weight decay of 0.001 for approximately 5,000 iterations to train the model on the THUMOS14 dataset. On ActivityNet-v1.2, the learning rate was set to 0.00003 for model optimization across 50,000 iterations. Additionally, $\lambda_1$, $\lambda_2$, $\mu_1$ were set sequentially to 1.50, 1.50, 1.00 (Detailed hyperparameter analysis can be found in the supplementary materials). The implementation of our model was based on PyTorch 1.8, and training was conducted on the Ubuntu 18.04 platform.

\subsection{Comparison with State-of-the-art Methods.}
In this section, we compare our proposed method with the current state-of-the-art temporal action localization methods, as presented in Tables \ref{table1} and \ref{table2}. For the THUMOS14 dataset, we selected DELU and Zhou et al.\cite{zhou2023improving} as our baseline methods. After incorporating our MLLM4WTAL, we achieved improvements of 1.8\% and 1.0\% in the average mAP (0.1:0.7) metric, respectively, surpassing current weakly supervised methods and even outperforming some fully supervised methods, such as SSN and P-GCN. On the larger-scale ActivityNet-v1.2 dataset, our method achieves a 0.5\% improvement in average mAP compared to existing state-of-the-art methods.

\subsection{Ablation Study}
To analyze the contribution of each of our proposed modules, we conducted extensive ablation experiments to evaluate our approach from multiple perspectives, as shown in  Table \ref{table3}.

% \begin{figure}[h]
% 	\centering
% 	\includegraphics[width=1.0\columnwidth]{sec/fig/MLLM-22-1.pdf}
% 	\caption{Bar Chart Analysis of Ablation Experiments.}
% 	\label{fig:hy}
%         \vspace{-5mm}
% \end{figure}

\noindent\textbf{Baseline.} In the baseline experimental setup, without using any of our proposed components, we achieved an average mAP of 37.4\%.

\noindent\textbf{Impact of KSM.} In the Baseline+KSM experimental setup, where we solely incorporated the proposed Key Semantic Matching (KSM), we achieved an average mAP of 38.9\%. The introduction of KSM resulted in a more significant mAP improvement, especially in high IoU settings. For instance, at IoU=0.7, there was a relative improvement of 1.6\% compared to the Baseline.

\begin{figure*}[t]
	\centering
	\includegraphics[width=\linewidth,height=3cm]{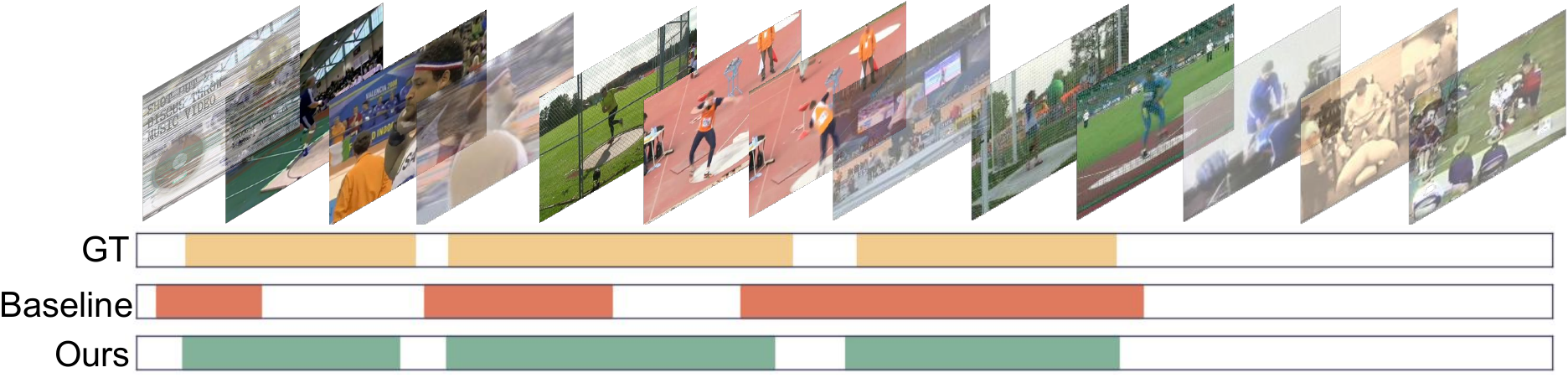}
        \includegraphics[width=\linewidth,height=3cm]{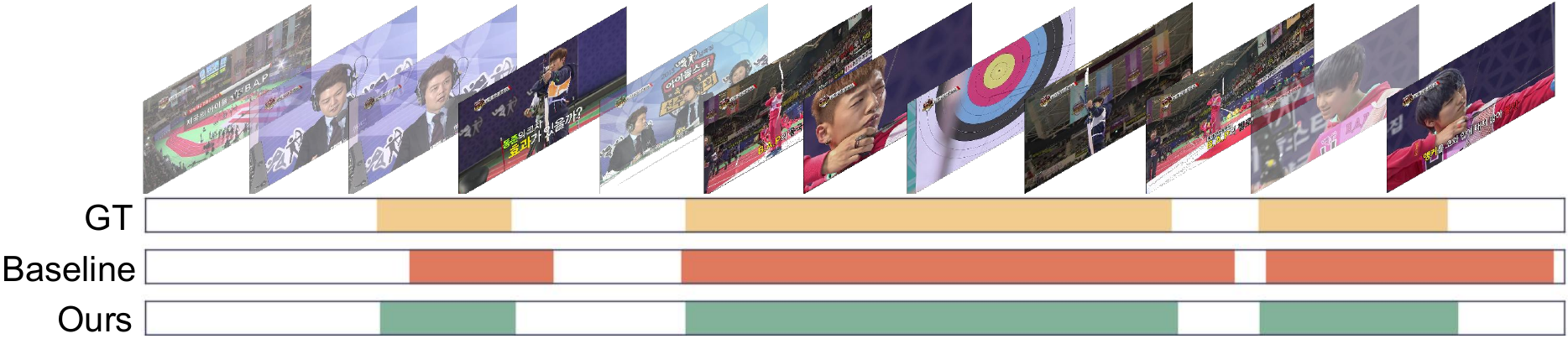}

	\caption{Visualization of ground-truth and predictions.}
	\label{fig:qua}
\end{figure*}

\noindent\textbf{Impact of CSR.} In the Baseline+CSR experimental setup, where only the proposed Complete Semantic Reconstruction (CSR) was utilized, we achieved an average mAP of 38.7\%. Similar to KSM, the introduction of CSR led to a more pronounced mAP improvement, with a relative increase of 2.2\% at IoU=0.7 compared to the Baseline.

\noindent\textbf{Combined Impact of KSM and CSR.} In the Baseline+KSM+CSR experimental setup, where both KSM and CSR were simultaneously employed, we achieved an average mAP of 39.2\%. The simultaneous introduction of KSM and CSR resulted in a more noticeable mAP improvement, particularly in high IoU settings. For example, at IoU=0.7, there was a relative improvement of 3.1\% compared to the Baseline.

\noindent\textbf{Impact of prior information.}
In Table \ref{ablation-mllm}, We analyzed the impact of MLLM from two perspectives: 
{(1)The impact of key semantic priors provided by MLLM.} We conducted an ablation study by replacing the key semantic priors provided by MLLM with the names of temporal action categories. As shown in Table \ref{ablation-mllm}, the average mAP  improved by 1.5\% when using key semantic priors.{(2)The impact of complete semantic priors provided by MLLM.} We conducted an ablation study by replacing the complete semantic priors provided by MLLM with fixed-template temporal action descriptions. As shown in Table \ref{ablation-mllm}, the average mAP improved by 1.6\% when using complete semantic priors.

\subsection{Generic validation}
It is important to emphasize the versatility of our MLLM4WTAL framework, as it can be plug-and-played into existing mainstream WTAL methods. To assess the generality of our framework, we systematically incorporated two classical WTAL methods, UM and CO2-Net, into our framework. Utilizing only UM resulted in an average mAP of 30.3\%, and subsequent integration of KSM and CSR led to average mAP improvements of 1.8\% and 2.3\%, respectively. Similarly, employing only CO2-Net yielded an average mAP of 36.0\%, with sequential addition of KSM and CSR resulting in average mAP enhancements of 1.9\% and 2.2\%, respectively. These findings robustly demonstrate the adaptability of our method in effectively enhancing various mainstream WTAL methods.
% \subsection{hyperparametric Analysis}

% \begin{figure}[ht]
%     \centering
%     % First row
%     \begin{subfigure}
%         \centering
%         \includegraphics[width=\linewidth]{sec/fig/MLLM-hp-1.pdf}
%         \caption{IMDP-rada-1-F1}
%         \label{fig:sub1}
%     \end{subfigure}
    
%     % Second row
%     \begin{subfigure}
%         \centering
%         \includegraphics[width=\linewidth]{sec/fig/MLLM-hp-2.pdf}
%         \caption{IMDP-rada-1-AUC}
%         \label{fig:sub3}
%     \end{subfigure}
%     \caption{The hyperparametric of our method.}
%     \label{2}
% \end{figure}

% \begin{figure}	
% 	\centering
% 	\begin{subfigure}[t]{1in}
% 		\centering
% 		\includegraphics[width=1in]{sec/fig/MLLM-hp-1.pdf}
% 		\caption{Caption 1}\label{fig:1a}		
% 	\end{subfigure}
% 	\quad
% 	\begin{subfigure}[t]{1in}
% 		\centering
% 		\includegraphics[width=1in]{sec/fig/MLLM-hp-2.pdf}
% 		\caption{Caption 2}\label{fig:1b}
% 	\end{subfigure}
% 	\caption{Main figure caption}\label{fig:1}
% \end{figure}

% We perform ablation experiments to investigate the effects of the hyperparameters $\lambda_1$ and $\lambda_2$ in the equilibrium loss $L_{KSM}$ and $L_{CSR}$, respectively. The results of these experiments are summarized in the figure above. Various values are tested to determine the optimal weights for each hyperparameter. Based on the experimental results, we empirically set $\lambda_1$ = 1.50 and $\lambda_2$ = 1.50 as they yield the best performance. These values were chosen to strike a balance between the two losses and achieve the best results for our framework.

\subsection{Qualitative Analysis}
As shown in Fig.\ref{fig:qua}, we present the visual results of temporal action localization. From the visualization, it can be observed that the baseline exhibits significant issues of over-complete and incomplete localization results. However, upon introducing our MLLM, which provides key semantic and complete semantic priors, the problems of over-complete and incomplete are noticeably alleviated.

%% file: sec/Conclusion.tex
\section{Conclusion}
We introduce a paradigm, MLLM4WTAL, to guide the Weakly-supervised Temporal Action Localization (WTAL) task. The MLLM4WTAL framework enhances the WTAL methods in two ways: Key Semantic Matching and Complete Semantic Reconstruction. Key Semantic Matching aims to utilize the key semantic priors provided by MLLM to activate crucial intervals of action instances. Complete Semantic Reconstruction aims to leverage the complete semantic priors provided by MLLM to mine as complete action instance intervals as possible. Integrating the prediction results of the above two modules, a dual prior interactive distillation strategy is employed to address the common issues of over-complete and incomplete in WTAL. Extensive experiments demonstrate that our approach achieves state-of-the-art performance on two popular datasets and can be plug-and-played into existing WTAL methods.

\section{Acknowledgement}
This work was supported by the National Key R\&D Program of China (2022YFB4701400/4701402), SSTIC Grant(KJZD20230923115106012,KJZD202309231149160 32,GJHZ20240218113604008), Beijing Key Lab of Networked Multimedia and National Natural Science Foundation of China under Grant 62202302.